\documentclass{article}
\usepackage{spconf,amsmath,graphicx,booktabs}
\usepackage[numbers,sort,compress]{natbib}
\usepackage{hyperref}

\newcommand{\etal}{\textit{et al.~}}
\renewcommand{\refname}{} 

\title{Depth from Defocus via Direct Optimization}
\name{Holly Jackson$^{\star}$ \qquad Caleb Adams$^{\dagger}$\qquad Ignacio Lopez-Francos$^{\dagger,\ddagger}$\qquad Benjamin Recht$^{\star}$}
\address{$^{\star}$ University of California, Berkeley, California, USA\\
$^{\dagger}$ NASA Ames Research Center, Moffett Field, California, USA\\
$^{\ddagger}$ KBR, Houston, Texas, USA}

\begin{document}
%
\maketitle
\begin{abstract}
Though there exists a reasonable forward model for blur based on optical physics, recovering depth from a collection of defocused images remains a computationally challenging optimization problem.  In this paper, we show that with contemporary optimization methods and reasonable computing resources, a global optimization approach to depth from defocus is feasible.  Our approach rests on alternating minimization.  When holding the depth map fixed, the forward model is linear with respect to the all-in-focus image.  When holding the all-in-focus image fixed, the depth at each pixel can be computed independently, enabling embarrassingly parallel computation.  We show that alternating between convex optimization and parallel grid search can effectively solve the depth-from-defocus problem at higher resolutions than current deep learning methods.  We demonstrate our approach on benchmark datasets with synthetic and real defocus blur and show promising results compared to prior approaches. Our code is available at \href{https://github.com/hollyjackson/dfd}{github.com/hollyjackson/dfd}.
\end{abstract}
\section{Introduction}
\label{sec:intro}

A century ago, Hermann von Helmholtz first noted that optical blur carries information about scene depth \cite{Helmholtz1924}.  This insight gave rise to a new 3D reconstruction problem:  given a set of images captured from a single viewpoint at different focus settings (i.e.~a focal stack), can we recover depth from the varying amounts of blur?  A clear forward model for defocus blur, which takes a depth map and in-focus image and produces a focal stack, has since been developed from optical physics approximations.  Theoretically, depth from defocus could be solved by simply finding the depth map and in-focus image that minimize the focal stack reconstruction error under the forward model.  However, the nonlinearity of the forward model has kept its direct inversion beyond the reach of standard optimization techniques in inverse problems.

Classically, engineers turned to other heuristics to quickly estimate depth maps.  Over the past decades, deep learning has outperformed such heuristics, but introduced a new reliance on training data which are expensive and difficult to acquire.  A handful of approaches have tried to attack the global optimization problem, but most do so indirectly – either by breaking it down into multiple smaller optimizations or by relying on regularization to generate a valid depth map.  

In this paper, we revisit a global optimization approach to depth from defocus.  We run a straightforward alternating minimization to solve for the two unknowns: the depth map and the all-in-focus (AIF) image.  A key observation – which, to our knowledge, has not been exploited in prior work – is that when the depth map is held fixed, the problem is linear in the AIF image.  In this case, the minimization step can be efficiently solved using convex optimization methods.  When the AIF image is held fixed, the complementary minimization for the depth map becomes massively parallelizable, amounting to nonlinear search at every pixel.  The optimization objective is simply the mean square reconstruction error of the focal stack.  Because this inverse problem is overdetermined (i.e.~there are more measured pixels than unknowns), regularization is not required to recover the depth and AIF image.

We demonstrate our approach on real images using both synthetic and real defocus blur.  Our method outperforms all state-of-the-art depth-from-defocus approaches on the NYUv2 and Make3D datasets when evaluated with synthetic defocus blur.  On real image blur data, qualitative evaluation indicates that our approach yields visually accurate and high-quality depth maps.  

Overall, we demonstrate that simple, direct optimization is not only feasible but outperforms more complex learning-based and regularized approaches.  This is achieved through three main contributions.  First, we identify and exploit the linear structure of the all-in-focus optimization subproblem, enabling efficient convex optimization.  Second, we show that the complementary depth optimization step is fully parallelizable.  Finally, our method exhibits iterative improvement against both supervised and self-supervised deep learning methods, as well as prior optimization-based methods, on synthetically-blurred images.

\section{Related work}
\label{sec:related-work}

Existing approaches can be divided into three categories: classical heuristic methods, global optimization methods (most closely related to our work), and deep learning-based approaches, which define the current state of the art.\medskip

\noindent \textbf{Classical depth from focus/defocus.} Classical approaches relied on heuristics rather than directly solving the underlying inverse problem.  Depth-from-focus (DFF) methods estimated a depth map by computing a per-pixel focus measure across each focus setting in the focal stack and selecting the setting that maximized sharpness \cite{Krotkov1988,DarrellWohn1988, BoveJr1993, NayarandNakagawa1994}.  By contrast, depth from defocus (DFD) estimated depth by measuring blur, rather than sharpness.  The most popular methods estimated the relative blur, i.e.~the blur needed to transform one defocused image into another \cite{Gokstorp1994, EnsandLawrence1991, Pentland1989, SubbaraoandSurya1994, WatanabeandNayar1998, ZiouandDeschenes2001}.  However, others attempted to blur all defocused images towards a common target \cite{XianandSubbarao2006}, or to deblur the defocused images \cite{Zhou2011}. No heuristic has proven to be universally optimal across scenes and imaging conditions; both DFF and DFD methods are sensitive to factors such as noise, scene texture, and blur kernel model \cite{Pertuz2013,MannanandLanger2016}.\medskip

\noindent \textbf{Global optimization.} Global optimization methods use the forward model to predict a focal stack for a candidate depth map, and then iteratively minimize the error between the predicted and observed focal stacks.  As formulated, this problem is nonconvex and can be ill-posed in texture-less regions, making the model difficult to invert directly.  As a result, most approaches separated the problem into multiple sequential optimization problems or added strong regularization terms to constrain the estimated depth map.
               
Moeller \etal \cite{Moeller2015} approach the depth-from-defocus problem as a variational energy minimization task, but the required regularization enforces artificial consistency across regions with distinct depth variations.  Favaro \etal \cite{Favaro2003} use iterative alternating minimization, similar to our approach, where the algorithm alternately updates the estimate for the depth map and the all-in-focus (AIF) image until convergence.  However, their AIF update is posed as a nonconvex deconvolution, whereas we reformulate this step as a simpler convex optimization.

Suwajanakorn \etal \cite{Suwajanakorn2015} undertake the more challenging problem of uncalibrated depth from defocus, jointly optimizing the depth map and camera parameters.  They use a complex, multi-stage pipeline with additional depth refinement and post-processing, as well as explicit regularization to bias the solution toward smooth depth estimates.  Our method borrows two key ideas from Suwajanakorn \etal \cite{Suwajanakorn2015}: AIF image stitching (which we use for initialization) and pre-computed “blur stacks” for fast forward-model evaluation.  Both are explained in detail in Section \ref{sec:optimization}.\medskip

\noindent \textbf{Deep learning.} From the late 2010s onward, researchers increasingly turned to deep learning-based methods to solve the depth-from-defocus inverse problem.  These methods rely on access to training datasets containing either real or synthetic focal stacks and additionally require ground-truth depth maps  \cite{Carvalho2018, Hazirbas2018, Maximov2020, Yang2022} and/or an AIF images \cite{GurandWolf2019, Lu2021, Wang2021} for supervision during training which are difficult and expensive to acquire.  Si \etal \cite{Si2023} propose a fully self-supervised approach that synthesizes focal stacks using the defocus forward model and optimizes a reconstruction loss with regularization.  Their method approaches the performance of supervised models and outperforms global optimization methods.

\section{Proposed method}\label{sec:methods}

\begin{figure}[t]
  \centering
   \includegraphics[width=0.95\linewidth]{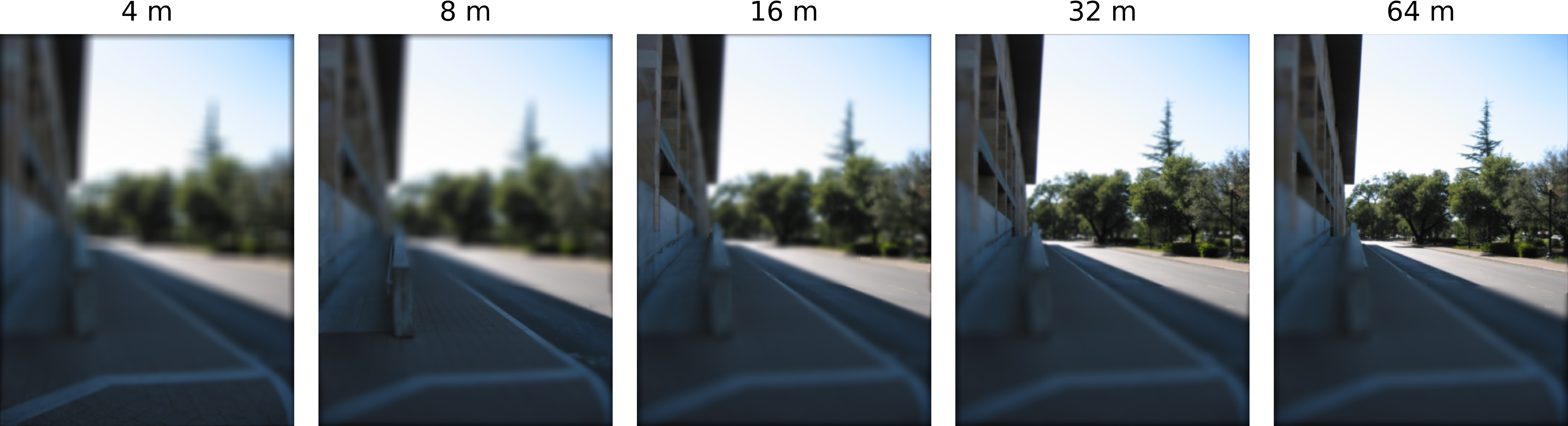}

   \caption{Example focal stack from Make3D \cite{Saxena2005,Saxena2009}, shown with synthetic defocus blur at five focus distances.}
   \label{fig:focal_stack_example}
\end{figure}

The problem we address in the depth-from-defocus inverse problem:  given an input focal stack (such as the example in Figure \ref{fig:focal_stack_example}), our goal is to recover both the underlying depth map and the all-in-focus (AIF) image.  The AIF image refers to the latent sharp image that would be obtained if all scene points were simultaneously in focus.  In Section \ref{sec:forward-model}, we first define the well-established depth-from-defocus forward model, which takes a depth map and an AIF image as input and produces a focal stack under specified camera settings.  We then solve the inverse problem using an alternating minimization scheme, enumerated in Section \ref{sec:optimization}. 

\subsection{Forward model for defocus blur}\label{sec:forward-model}

\begin{figure}[t]
  \centering
   \includegraphics[width=0.95\linewidth]{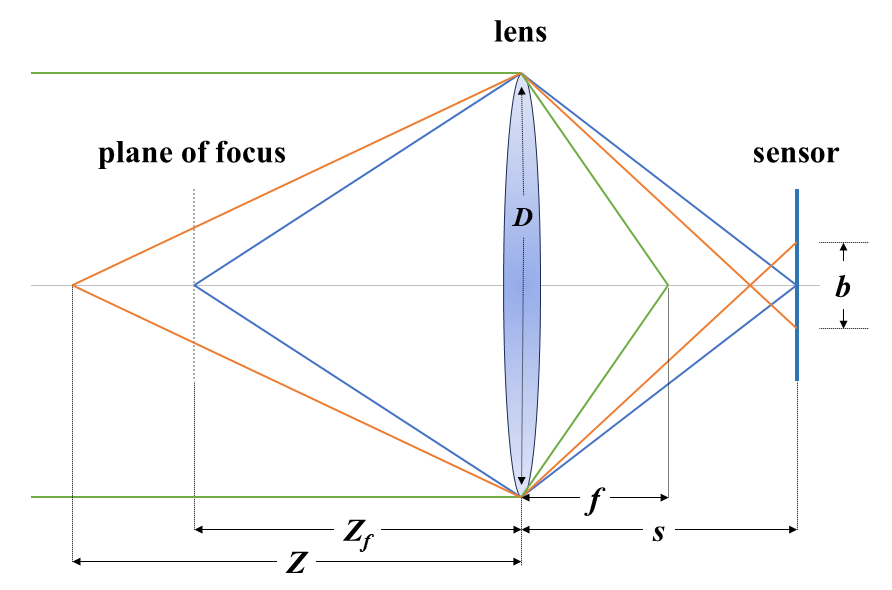}
\caption{Illustration of a thin lens with aperture diameter $D$ and focal length $f$.  Parallel rays from infinity focus at $f$ (green).  A point at the focus distance ($Z_f$) forms a sharp image on the sensor at distance $s$ (blue).  Points away from the focus plane form a circle of confusion with blur diameter $b$ (orange).}
   \label{fig:thinlens}
\end{figure}

Given a depth map $Z$ and an all-in-focus image $I$, we can model the captured focal stack under different camera settings, such as varying focus distance or aperture diameter.  The forward model for defocus blur follows directly from the thin lens law and basic trigonometry (Figure \ref{fig:thinlens}).

Each point on an object being imaged (or, equivalently, each pixel in the AIF image) lies either in front of, behind, or on the plane of focus.  A point lying on the plane of focus (at $Z_f$) projects to a single point on the image sensor, producing a perfectly sharp pixel.  However, when a scene point is not on the plane of focus, the rays passing through the lens converge either in front of or behind the image sensor.  This will blur the radiance of the corresponding scene point across a small area on the image plane, an effect often called a blur circle or circle of confusion (marked by its diameter $b$ in Figure \ref{fig:thinlens}).  We can express the diameter of the blur circle $b$ in terms of the camera parameters and the scene depth at a given point.

For an image captured using a lens with aperture diameter $D$ and focal length $f$ at a focus distance $Z_f$, the resulting blur diameter $b$ for an object point at depth $Z$ can be expressed:

\begin{equation}
    b = D \frac{\lvert Z - Z_f \rvert}{Z} \frac{f}{Z_f - f}
\end{equation}

\noindent The absolute value ensures that $b$ is sign-agnostic to the side of the focal plane.  The blur radius in pixels can be computed $\sigma = b / (2 \cdot ps)$, where $ps$ is the pixel size.

We approximate the effect of defocus by convolving the AIF image with a spatially-varying Gaussian kernel, parameterized by $\sigma$.  Equation \ref{eq:gaussian} shows the construction of a Gaussian kernel $G_{i,j}$ centered around a given pixel $(i,j)$.  Here, $u$ and $v$ denote the offsets from the kernel center.

\begin{equation}
	G_{i,j}(u,v) = \exp\left(-\frac{u^2+v^2}{2\sigma^2}\right)
    \label{eq:gaussian}
\end{equation}

\noindent The normalization factor $\frac{1}{2\pi\sigma^2}$ is omitted, as each Gaussian kernel is normalized post-construction.  Following prior work \cite{GurandWolf2019,Si2023}, we crop the kernel to a fixed size, which ignores negligible diffuse effects at the edges and improves computational efficiency.   We allow the maximum kernel size to scale with the input image dimensions up to a specified upper limit.  We also impose a minimum blur threshold in the forward model for numerical stability such as in \cite{Si2023}.

For each pixel, a unique Gaussian kernel is constructed and applied locally to the AIF image $I$.  The result is a defocused image $J$.  Because each kernel operates only at its corresponding pixel location, the convolution simplifies to a pointwise product.  Accordingly, each pixel $(i,j)$ in the defocused image can be written as:

\begin{equation}
	J_{i,j} = \sum\limits_{u, v} G_{i,j}(u,v) \cdot I_{i+u, j+v}
\end{equation}

\noindent This suggests the entire forward model can be reformulated as a sparse matrix multiplication $AI=J$.  For a focal stack, we just stack these sparse operators for each focus setting $Z_f$.

\subsection{Direct optimization of the inverse problem}\label{sec:optimization}

\noindent \textbf{Initialization.} To begin, one can initialize a guess for either the depth map or the AIF image.  In practice, we initialize the AIF image using the stitching algorithm proposed by Suwajanakorn \etal \cite{Suwajanakorn2015}.  Following \cite{Suwajanakorn2015}, we formulate the problem as a multi-label Markov Random Field optimization problem over a grid.  We minimize a weighted two-term energy, where one term penalizes blur and the other term encouraging stitching from the same or closely-neighboring focal stack layers.  The first term is computed as the sum of a per-pixel sharpness measure over a small Gaussian-weighted patch centered at each pixel.  In our implementation, we use the Sobel gradient magnitude as the sharpness measure.  The second term is defined as the absolute difference between adjacent pixels' indices in the focal stack.\medskip

\noindent \textbf{Alternating minimization.} The optimization proceeds via an alternating minimization scheme with two main stages: (1) hold the AIF image fixed and optimize for the depth map; and (2) hold the depth map fixed and optimize for the AIF image.  At each iteration, we apply the forward model to generate a predicted focal stack based on the current estimates of the AIF image and depth map. The reconstruction error, serving as the optimization objective, is computed as the mean squared error (MSE) between the predicted and input focal stacks.  The alternating updates are performed for a fixed number of epochs, or until convergence to a desired MSE.  The method requires input of camera parameters, including focal length, aperture diameter, and the focus distances used to capture the focal stack, as well as a specified depth range for the scene.\medskip

\noindent \textbf{Solving for the depth map.} When the AIF image is fixed, the optimization proceeds in two stages.  First, we perform a grid search over the input depth range, sampling $n=100$ candidate depths to identify the value that minimizes the reconstruction loss at each pixel.  Naively, we could evaluate the forward model on $n$ depth maps, each set uniformly to a candidate depth. However, since every pixel would be blurred by the same kernel, we can instead run $n$ fast convolutions, building a ``blur stack'' following Suwajanakorn \etal \cite{Suwajanakorn2015}. We compute the per-pixel MSE at leach layer in the blur stack and select the best candidate depth per pixel. Since these choices are independent, this yields the same result as explicitly evaluating the forward model at all $n$ candidate depths.

During grid search, we can replace the per-pixel MSE with a windowed MSE over a local patch, so each pixel selects the depth that best explains its neighborhood rather than just itself, encouraging locally-smooth depth. Depth is still chosen independently at each pixel, so the search remains fully parallelizable. This is done for two of the three datasets we evaluate in Section \ref{sec:results}, with a user-specified window size.

Second, we refine the grid search estimate using a golden-section search over a narrow window initialized around the minimum grid point at each pixel.  We set a maximum iteration limit of 100 and a convergence tolerance of $10^{-2}$.  Finally, we perform a pixel-wise comparison between the current and previous depth maps, selecting the value that minimizes the MSE for each pixel.  This guarantees monotonically-decreasing loss as the alternating minimization progresses.\medskip

\noindent\textbf{Solving for the AIF image.} When the depth map is fixed, we solve for the AIF image using Nesterov's accelerated gradient method, specifically following the Fast Iterative Shrinkage-Thresholding Algorithm (FISTA) described in Section 4 of Beck and Teboulle \cite{BeckTeboulle2009}.  During each FISTA update, the AIF reconstruction is clipped to a valid range (e.g.~$[0, 255]$).  The step size is set to the reciprocal of the Lipschitz constant of the gradient, as recommended in the paper \cite{BeckTeboulle2009}.  For efficiency, we approximate the Lipshitz constant by power iterations.  We run the FISTA updates for a fixed number of iterations $T_0$, which increases progressively by a user-defined multiplier $\alpha > 1$ as the alternating minimization progresses.  In our experiments, we used either of two sets of parameter values in our experiments:  $T_0 = 200$ with $\alpha = 1.05$, or a more aggressive setting of $T_0 = 10$ with $\alpha = 2$.

\begin{figure}[t]
  \centering
  \includegraphics[width=\linewidth]{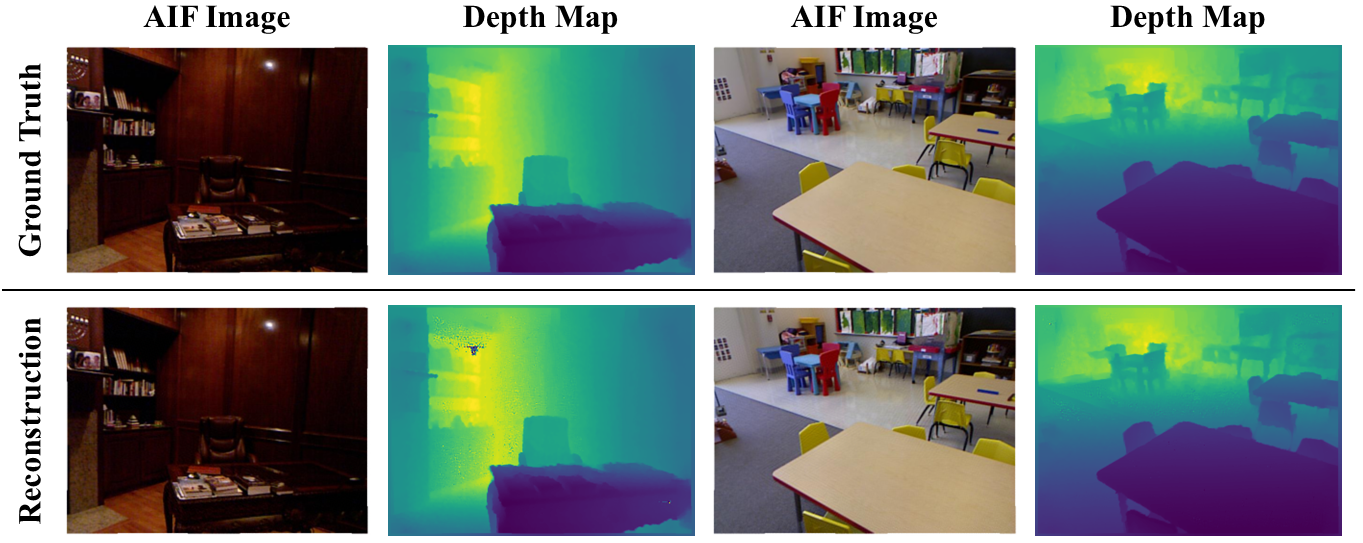}
  \caption{Qualitative results on two NYUv2 examples.  Top: ground-truth AIF images and depth maps.  Bottom: reconstructed AIF images and depth maps.  RMSEs (left to right) are 0.0731 and 0.0230.}
  \label{fig:NYUv2_examples}
\end{figure}

\section{Experiments}

This section describes the datasets, their configurations, and the experimental results.  All experiments were run entirely on CPU on a high-performance compute system with 1 TB of RAM and 72 physical cores, allowing dozens of focal stacks to be analyzed in parallel.

\subsection{Datasets}\label{sec:datasets}

We evaluate our method on three datasets.  Two contain real images that we synthetically blur to form focal stacks.  The third consists of real images captured at different focus settings with naturally-varying blur.\medskip

\noindent \textbf{NYUv2.} NYUv2 contains 1,449 samples with ground-truth depth maps and all-in-focus images \cite{Silberman2012}, and has been used in prior depth-from-defocus work \cite{GurandWolf2019,Si2023,Lu2021}. We generate synthetic focal stacks using our forward model with the same camera parameters as Si \etal \cite{Si2023}: focal length $f = 50$ mm, aperture diameter $D = f / 8$, focus distances $Z_f = [1, 1.5, 2.5, 4, 6]$ m, and pixel size $ps = 1.2 \times 10^{-5}$ m/px.  The input depth range of the dataset is $[0.1, 10]$ m.  Each example was run for 40 epochs with $T_0 = 200$ and $\alpha = 1.05$ and blur kernels with radius under 2 pixels are clipped.\medskip

\noindent\textbf{Make3D.} Make3D contains 533 images with ground-truth depth maps \cite{Saxena2005,Saxena2009}.  Following prior work \cite{Saxena2005,Saxena2009,GurandWolf2019}, images are resized to $460\times 345$.  We use a depth range of $[0.01, 80]$ m, extract camera parameters from EXIF metadata, and estimate pixel size manually.  Synthetic focal stacks are generated at seven focus distances $[1, 2, 4, 8, 16, 32, 64]$ m.  Blur radii under 0.5 pixels are clipped.  Each example was run for 5 epochs with $T_0 = 10$ and $\alpha = 2$.  We use windowed MSE during grid search with a window size of 5.\medskip

\noindent \textbf{Mobile phone focal stacks.} This dataset contains 11 focal stacks captured on a Samsung Galaxy S3 during autofocus \cite{Suwajanakorn2015}, each with 14–33 images and no ground-truth depth.  We use the authors' pre-aligned dataset and calibrated camera parameters.  Depth ranges are set to $[1,800]$ (pixels) with $n=200$ grid search samples.  Blur kernels with radius $< 0.1$ are clipped.  Each example was run for 5 epochs with $T_0 = 10$ and $\alpha = 2$.  Evaluation is qualitative.  We use windowed MSE during grid search with a window size of 50; because the scenes are captured at close range, this window typically lies within a locally constant-depth region.

\subsection{Results}\label{sec:results}

Evaluation follows the same metrics in prior work, including the root mean squared error (RMSE) and the absolute relative error (AbsRel) between the ground-truth and reconstructed depth maps.  For NYUv2, we also report the $\delta_1$, $\delta_2$, and $\delta_3$ depth accuracy metrics \cite{Eigen2014}, which measure the percentage of pixels within factors of 1.25, $1.25^2$, and $1.25^3$ of the ground-truth depth, respectively.  For consistency with previous learning work, we report our results over the test set, though the performance of our method was consistent over the training and test sets (full results in Supplementary Materials). \medskip

\noindent \textbf{NYUv2.} Table \ref{tab:NYUv2_results} presents a comparison of our results on NYUv2 against state-of-the-art depth-from-defocus methods.  Our method achieves the lowest RMSE and AbsRel and highest accuracy metrics ($\delta_1, \delta_2, \delta_3$) among all prior depth-from-defocus approaches, including both analytical and deep learning methods -- even those supervised on ground-truth depth maps.  Our method also outperforms all monocular depth estimation methods on NYUv2 documented by Gur and Wolf \cite{GurandWolf2019} (see Supplementary Materials for details).

Our qualitative results (Figure \ref{fig:NYUv2_examples}) show high fidelity to the ground-truth depth maps, preserving fine details and avoiding the oversmoothing present in other approaches as a result of heavy regularization.  Occasional, well-contained artifacts appear in low-texture regions.  \medskip

\begin{figure}[t]
  \centering
  \includegraphics[width=\linewidth]{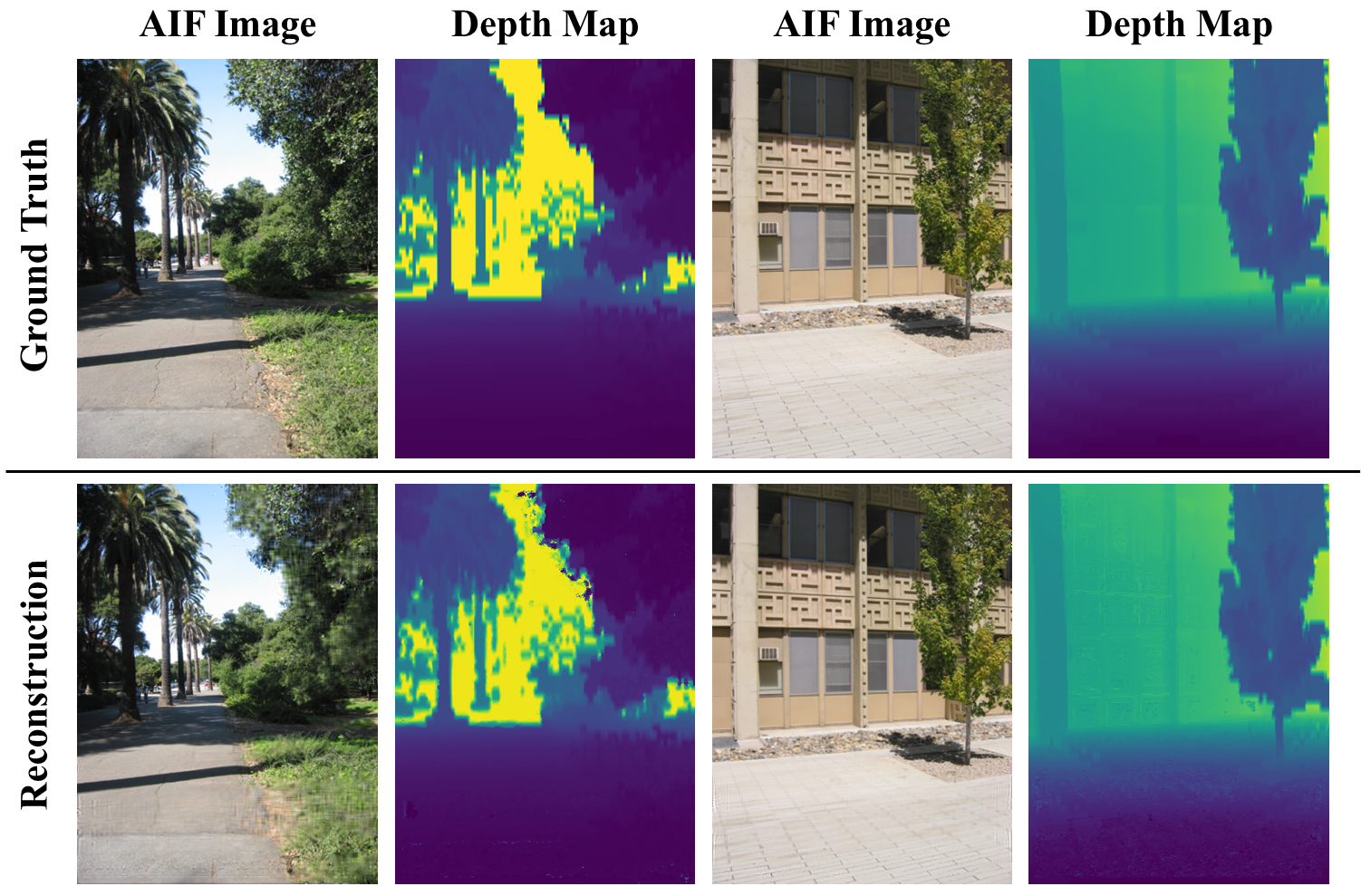}
  \caption{Qualitative results on two Make3D examples.  Top: ground-truth AIF images and resized depth maps.  Bottom: reconstructed AIF images and depth maps. C2 RMSEs (left to right) are 5.439 and 0.243.}
  \label{fig:Make3D_examples}
\end{figure}

\noindent \textbf{Make3D.} We evaluate on Make3D following prior work, reporting errors under two standard depth ranges: C1 (0–70 m) and C2 (0–80 m) \cite{Saxena2005, Saxena2009}.  The only prior depth-from-defocus method evaluated on Make3D is Gur and Wolf \cite{GurandWolf2019}, which uses focal stacks during training but predicts depth from a single defocused image at test time.  Our method outperforms this partially-supervised approach under both error ranges (Table \ref{tab:Make3D_results}) and exceeds all monocular depth estimation methods reported in \cite{GurandWolf2019} (see Supplementary Materials). Qualitative results are shown in Figure \ref{fig:Make3D_examples}.\medskip

\begin{figure}[t]
  \centering  
  \includegraphics[width=\linewidth]{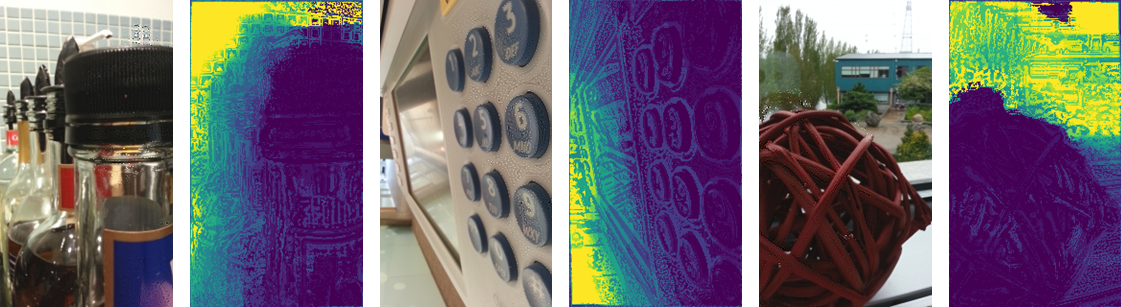}
  \caption{Reconstructed AIF images (left) and depth maps (right) for three scenes from the mobile phone dataset \cite{Suwajanakorn2015}.}
\label{fig:mobile_depth}
\end{figure}

\noindent \textbf{Mobile phone focal stacks.} Since ground-truth depth maps are not available for the mobile phone focal stacks \cite{Suwajanakorn2015}, we perform a qualitative evaluation.  Figure \ref{fig:mobile_depth} presents our results on representative images from the dataset, with predicted depth maps normalized across the depth range.  Qualitatively, our results are comparable to recent state-of-the-art methods (see Figure 5 in Si \etal \cite{Si2023}).

\begin{table}[t]
\centering
\setlength{\tabcolsep}{3pt} 
\small
\begin{tabular}{lccccc}
\toprule
Method & RMSE$\downarrow$ & AbsRel$\downarrow$ & $\delta_1\uparrow$ & $\delta_2\uparrow$ & $\delta_3\uparrow$\\
\midrule
\multicolumn{6}{c}{\textit{Supervised learning}} \\
{DFF-FV \cite{Yang2022}}    & 0.285 & {0.470}   & 0.956 & 0.979   & 0.988   \\
{DFF-DFV \cite{Yang2022}}   & 0.232 & {0.445}   & 0.967                         & 0.980   & 0.990   \\
\midrule
\multicolumn{6}{c}{\textit{Self-supervised w/ AIF}} \\
{Gur \& Wolf \cite{GurandWolf2019}}             & 0.649 & {0.184}   & 0.720                         & 0.887   & 0.951   \\
{Defocus-Net \cite{Lu2021}}    & 0.623 & {0.176}   & 0.732                         & 0.887   & 0.951   \\
{Focus-Net \cite{Lu2021}}    & 0.611 & {0.172}   & 0.748                         & 0.892   & 0.949   \\
\midrule
\multicolumn{6}{c}{\textit{Self-supervised}} \\
{Si \etal \cite{Si2023}}         & 0.325 & {0.170}   & 0.950                         & 0.979   & 0.987   \\
\midrule
\multicolumn{6}{c}{\textit{Analytical}} \\
{Moeller \etal \cite{Moeller2015}}      & 0.985 & {0.263}   & 0.670                         & 0.778   & 0.912   \\
{Suwajanakorn \etal \cite{Suwajanakorn2015}}    & 0.950 & {0.250}   & 0.688                         & 0.802   & 0.917   \\
{Ours}                & \textbf{0.109} & \textbf{8.37e-3} & \textbf{0.992}                         & \textbf{0.994}   & \textbf{0.995}   \\
\bottomrule
\end{tabular}
\caption{Comparison of depth-from-defocus results on the NYUv2 test set \cite{Silberman2012}. Results for \cite{Moeller2015,Suwajanakorn2015,GurandWolf2019,Lu2021} are taken from \cite{Lu2021}, and results for \cite{Si2023,Yang2022} from \cite{Si2023}. Lower is better for RMSE and AbsRel; higher is better for $\delta_1$, $\delta_2$, and $\delta_3$.}
\label{tab:NYUv2_results}
\end{table}

\begin{table}[t]
\centering
\setlength{\tabcolsep}{3pt} 
\small
\begin{tabular}{lcccc}
\toprule
& \multicolumn{2}{c}{C1} & \multicolumn{2}{c}{C2}\\
Method & RMSE$\downarrow$ & AbsRel$\downarrow$ & RMSE$\downarrow$ & AbsRel$\downarrow$\\
\midrule
\multicolumn{5}{c}{\textit{Self-supervised w/ AIF}} \\
{Gur \& Wolf \cite{GurandWolf2019}}            & 7.474   &   0.262 & 9.248 & {0.269}    \\
\midrule
\multicolumn{5}{c}{\textit{Analytical}} \\
{Ours}                   &  \textbf{2.277} & \textbf{0.0663} & \textbf{5.630} & \textbf{0.0695}\\ 
\bottomrule
\end{tabular}
\caption{Comparison of depth-from-defocus results on the Make3D test set.  For \cite{GurandWolf2019}, we report their best RMSE (F6, rendered focal stack with six focus settings). Lower is better.}
\label{tab:Make3D_results}
\end{table}

\section{Conclusion}

In this paper, we proposed a simple direct optimization approach for depth from defocus. Despite its simplicity, our direct optimization method outperforms sophisticated deep learning and complex nonlinear optimization techniques. However, our approach still suffers from limitations shared by across depth-from-defocus methods, such as poor performance in low-texture regions. While smoothness regularization in other methods can sometimes suppress this issue at the cost of fine detail, our method may instead produce localized artifacts. These artifacts can be effectively addressed using an optional post-processing step described in the Supplementary Materials. Our method also assumes known camera calibration parameters, which can be restrictive in some real-world applications (e.g.~mobile phones). To alleviate this, we plan to investigate methods for blind calibration in future work. Along similar lines, our approach assumes the camera can be well approximated by an ideal thin-lens model, and finding optimization methods to adapt to different lens models could be a promising future direction. Performance is also sensitive to the choice of focus settings; if the chosen focus distances do not produce sufficiently distinct images in the focal stack, the defocus cues become ambiguous.  Finally, the forward model and sparse matrix formulation scale quadratically with image size, and we aim to address this limitation through a fully GPU-optimized implementation in future work.

\vfill\pagebreak

\noindent\begin{center}
\bfseries\MakeUppercase{REFERENCES}
\end{center}

\makeatletter
\renewcommand{\refname}{\vspace*{-1.8\baselineskip}}
\makeatother

{\small
\def\IEEEbibitemsep{0pt} 
\bibliographystyle{IEEEbib}
\bibliography{refs}
}

\end{document}


%

\begin{center}
{\large \bfseries \MakeUppercase{Supplementary Materials}\par}
\end{center}
\vskip 1.5em

\section{Overview}

In the supplementary materials, we provide additional details on the method and experiments that were omitted from the main paper for space.  In Section \ref{sec:supp-method}, we expand on the method.  We describe how we construct the sparse forward operator $A$, detail the windowed MSE which can be used during grid search, and propose a post-processing method which could remove artifacts in our results.  In Section \ref{sec:additional-results-supp}, we present extended experimental results. This includes additional qualitative results on NYUv2 \cite{Silberman2012} and Make3D \cite{Saxena2005, Saxena2009}, and full qualitative results on the mobile phone focal stacks \cite{Suwajanakorn2015}. We also include convergence analysis plots and additional quantitative comparisons to monocular depth estimation methods.  Finally, in Section \ref{sec:supp-performance}, we discuss performance tradeoffs, including Make3D results without windowed MSE and NYUv2 results under different iteration budgets.

\section{Additional Method Details}\label{sec:supp-method}

Below we expand on specific aspects of our method.  In Section \ref{sec:supp-A}, we describe explicitly how we construct the sparse operator $A$, i.e.~the matrix that explicitly encodes the forward model used to generate defocused images from an AIF and depth estimate.  We then go over the details of the windowed mean squared error, which was enabled for grid search on Make3D and the mobile phone focal stacks (Sec. \ref{sec:supp-windowed-mse}).  Finally, in Section \ref{sec:supp-post-processing}, we propose an optional post-processing step to remove artifacts, which was mentioned in the limitations section (Sec. 4.3) of the main paper.

\subsection{Constructing the forward operator $A$}\label{sec:supp-A}

To make Equation 3 in the main paper fully explicit, we define the linear operator $A$ that maps the all-in-focus image $I$ to a defocused image $J$ for a given focus setting. As described in the main paper, for each pixel $(i,j)$ we construct a pixel-specific Gaussian kernel $G_{i,j}$ and apply it locally to $I$, so that the defocused value $J_{i,j}$ is a weighted sum of intensities in a neighborhood around $(i,j)$. The full forward model can be expressed compactly as a sparse matrix multiplication $AI=J$.  Repeating this computation for each focus setting yields the full focal stack.

\begin{figure}
  \centering
  \includegraphics[width=0.7\linewidth]{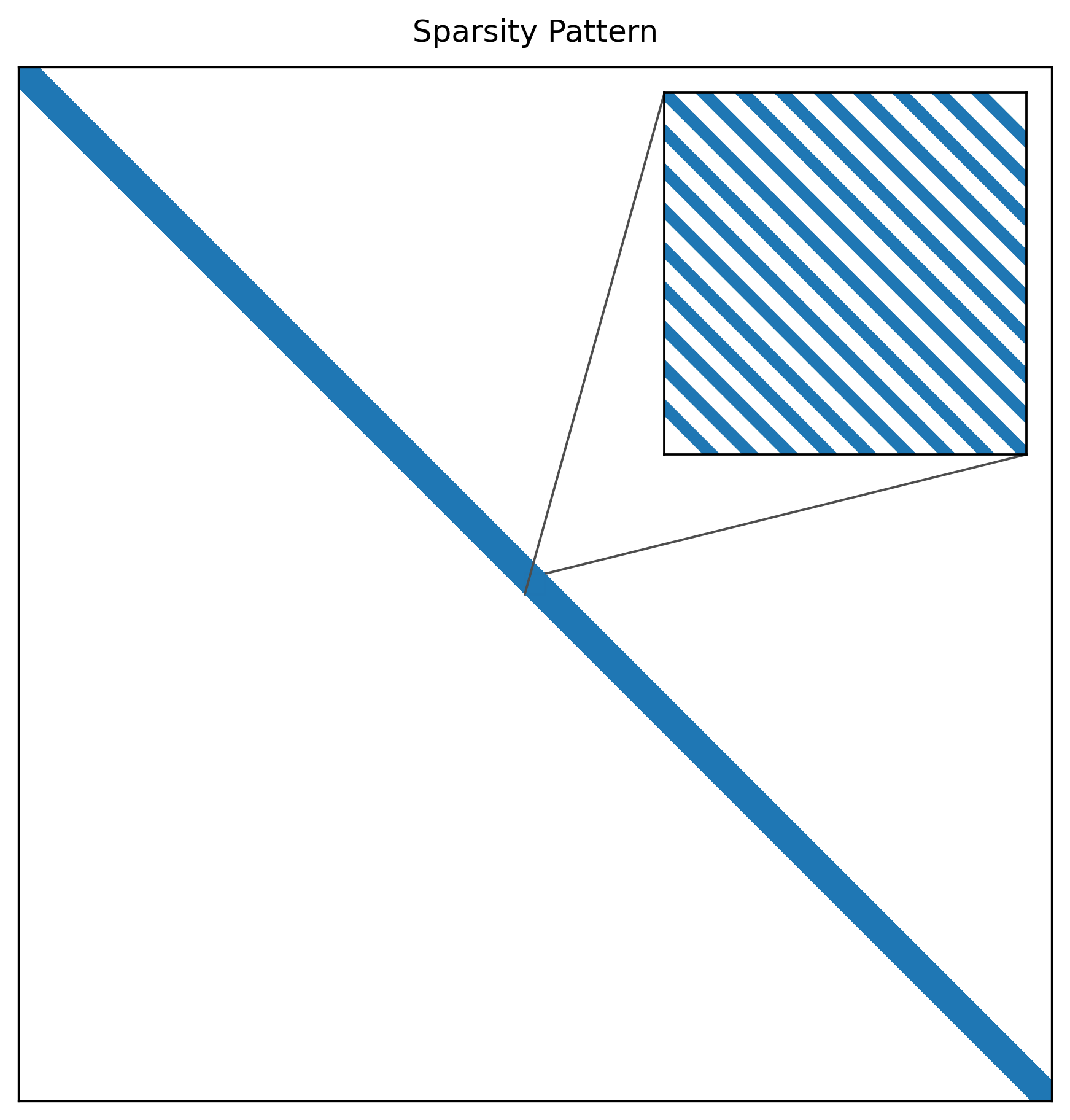}
  \caption{Sparsity pattern of $A$, which is constructed based on the depth map $Z$ and camera parameters.  The example shown is for the Make3D data, where the (resized) images are 460 $\times$ 345.  Here the maximum blur kernel size is set to 17 $\times$ 17.}
  \label{fig:Axb}
\end{figure}

Let's explicitly define $I$ as the the $m \times n$ all-in-focus image (flattened to be a length-$mn$ vector), $J$ as the $m \times n$ defocused result (flattened to be a length-$mn$ vector) for a specific focus setting, and $A$ as a large but sparse $mn \times mn$ matrix. This matrix encodes the blur kernels defined at each $G_{i,j}$.  Its sparsity pattern is illustrated in Figure \ref{fig:Axb}. By constructing $A$ appropriately, each defocused image in the focal stack can be efficiently computed in one step.

Let the depth map $Z$ and all-in-focus image $I$ be vectorized with flattened pixel index:
$$p = im + j, \qquad i \in \{0, \dots, m-1\}, j \in \{0, \dots, n-1\}.$$

For each output pixel $(i,j)$, the forward model applies a local blur kernel 
$G_{i,j}$ according to Equation 3 of the main paper.  Setting the maximum kernel size \texttt{max\_kernel\_size}, we can set radius 
$r=\lfloor \texttt{max\_kernel\_size}/2\rfloor$ and write the defocused pixel value as follows:

\begin{equation}
J_{i,j} = \sum\limits_{u = -r}^r\sum\limits_{v=-r}^r G_{i,j}(u,v) \cdot I_{i+u,j+v}
\end{equation}

\noindent where terms that fall outside the image bounds are omitted.

We encode this operation with a sparse matrix $A$.  We can define each row $p$ corresponding to pixel $(i,j)$ as follows.  Specifically, for each offset $(u,v)\in[-r,r]^2$ such that $(i+u,j+v)$, we set:

\begin{equation}
    A_{p,q} = G_{i,j}(u,v), \qquad q = (i+u)m + (j+v)
\end{equation}

\noindent All other entries in row $p$ are zero.

Each defocused image in the focal stack can be computed efficiently in a single matrix multiplication step once the sparse matrix $A$ is constructed. This process is highly efficient, producing the forward model results almost instantaneously in practice.

\subsection{Windowed mean squared error for grid search}\label{sec:supp-windowed-mse}

To obtain locally-smooth results, we enable a windowed MSE during grid search on Make3D and the mobile phone focal stacks. For each pixel, we assume depth is approximately constant within a local neighborhood and compute the reconstruction error over that window rather than at a single pixel. The windowed error replaces the standard per-pixel MSE, but depth is still selected independently at each pixel, so the search remains fully parallelizable.  Below, we explain the computation of the windowed MSE in detail.

For a predicted focal stack $\hat{\mathbf{J}} = (\hat{J_1}, \hat{J_2}, \dots, \hat{J_K})$ and an observed focal stack $\mathbf{J} = (J_1, J_2, \dots, J_K)$, we first compute the standard per-pixel mean squared reconstruction error by averaging squared differences over the focus settings $K$ and color channels $C$:

\begin{equation}
    e(i,j) = \frac{1}{KC} \sum\limits_{k, c} (J_k(i,j,c) - \hat{J_k}(i,j,c))^2
\end{equation}

\noindent This produces a 2D error map $e \in \mathbb{R}^{m \times n}$.

To encourage spatially smoother depth selections during grid search, we can replace this pointwise error with a windowed (locally averaged) error. For each pixel $(i,j)$, we define a square window $W_r(i,j)$ of radius $r = \lfloor \texttt{window\_size} /2 \rfloor$, and compute:

\begin{equation}
    \tilde{e}(i,j) = \frac{1}{\lvert W_r(i,j)\rvert} \sum\limits_{(u,v) \in W_r(i,j)} e(i+u,j+v)
\end{equation}

\noindent where $W_r(i,j)$ is clipped at image boundaries and 
$\lvert W_r(i,j)\rvert$ denotes the number of valid pixels contributing to the sum.  This windowed error can then be used as the objective for per-pixel depth selection.

\begin{figure}[t]
  \centering  
  \includegraphics[width=0.9\linewidth]{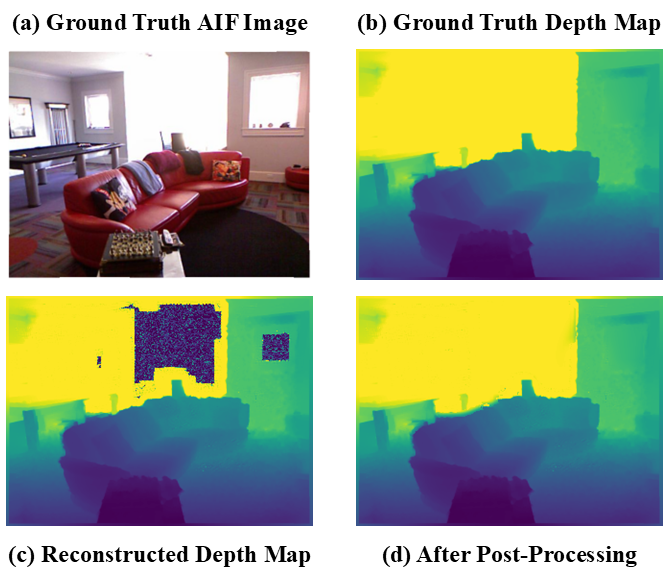}
  \caption{Example NYUv2 result with one of the highest RMSEs. (a) Ground-truth AIF image. (b) Ground-truth depth. (c) Reconstructed depth (RMSE = 1.397). (d) After post-processing (TV threshold = 0.4), reducing RMSE to 0.0604.}
\label{fig:limitations}
\end{figure}

\subsection{Optional post-processing}\label{sec:supp-post-processing}

Optionally, we could apply a post-processing step to the final depth map to eliminate artifacts, though this was not used in any of our experiments in the main paper.  We identify regions of the depth map that contain artifacts using a chosen metric, and pixels located within these regions are replaced with the mean of valid neighboring depths.  This strategy is motivated by the observation that most artifacts occur in low-texture regions (e.g.~a uniform-color wall), where all depth hypotheses produce similar local focal stacks.  We identify two possible metrics to detect artifacts.  One is to identify uniform regions in the AIF image, although this approach tends to be overly aggressive. However, in practice, we found it is more effective to identify regions of the depth map exhibiting high total variation, determined by a user-defined threshold.   These correspond to areas where the reconstruction is highly noisy, which often occurs in low-texture regions of the AIF image.

Figure \ref{fig:limitations} shows an example with the one of the highest (worst) RMSE on the NYUv2 dataset, illustrating the characteristic artifacts.  Note the large region of white wall and small region of bright window in the AIF image (\ref{fig:limitations}a) are reconstructed incorrectly by our method (\ref{fig:limitations}c).  However, the optional post-processing step can easily resolve such artifacts when they appear (\ref{fig:limitations}d), without oversmoothing fine details in the reconstruction.

\begin{figure}[t]
  \centering
   \includegraphics[width=\linewidth]{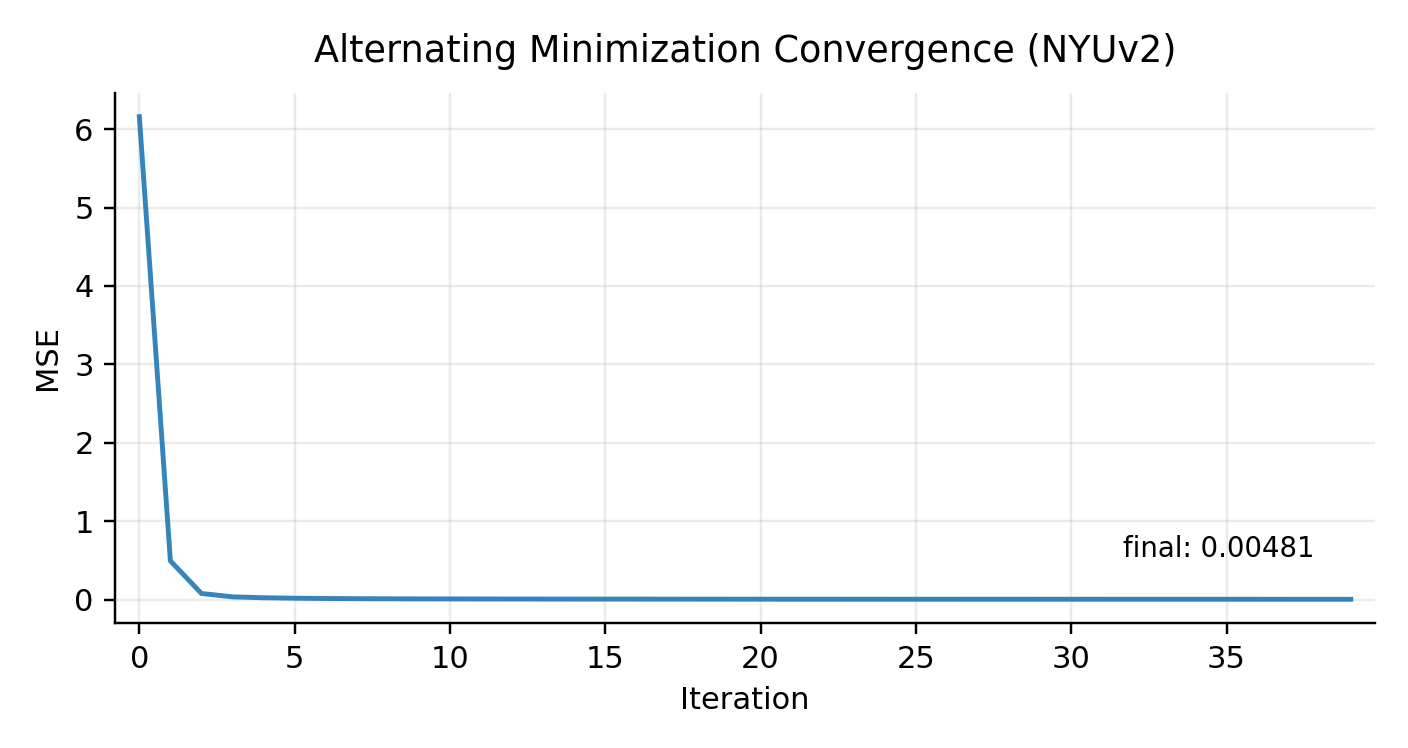}
  \caption{Loss vs. alternating-minimization iteration for NYUv2 main-paper experiment (run for 40 epochs).}
  \label{fig:convergence-nyuv2}
\end{figure}

\begin{figure}[t]
  \centering
   \includegraphics[width=\linewidth]{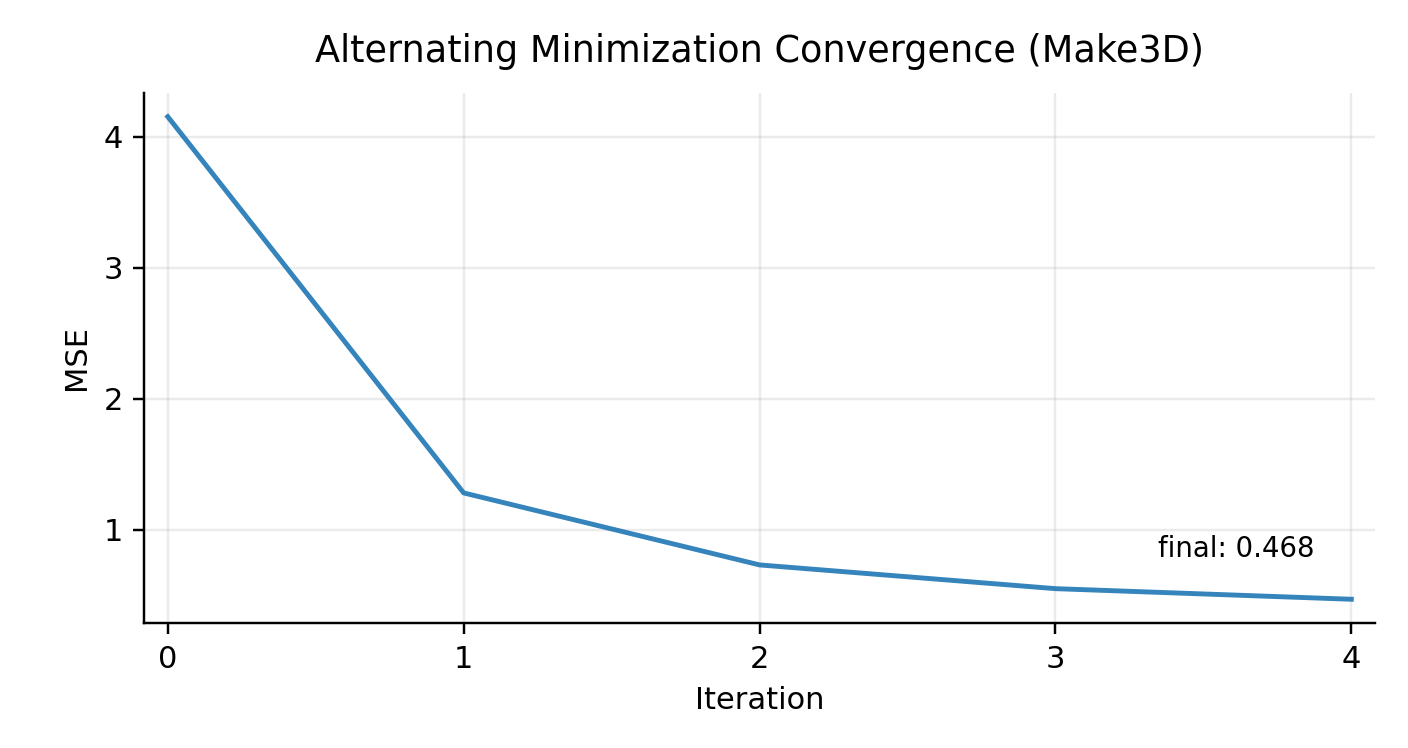}
  \caption{Loss vs. alternating-minimization iteration for Make3D main-paper experiment (run for 5 epochs).}
  \label{fig:convergence-make3d}
\end{figure}

\section{Extended experimental results}
\label{sec:additional-results-supp}

Below we expand upon our experimental results presented in the main paper.  In Section \ref{subsec:metrics-supp}, we provide an in-depth summary and definitions of metrics used for quantitative analysis.  In Section \ref{subsec:qual}, we show extended qualitative results on the NYUv2 \cite{Silberman2012} and Make3D \cite{Saxena2005, Saxena2009} datasets.  In Section \ref{subsec:mono-depth-comp}, we present extended comparison tables that span both depth-from-defocus and monocular depth estimation methods.  This broader evaluation highlights the strength of our framework as a general approach to depth estimation from a single viewpoint.  Finally, in Section \ref{subsec:mobile-depth-supp}, we present the full results on the mobile-phone focal stacks from Suwajanakorn \etal \cite{Suwajanakorn2015}.

\subsection{Metrics}
\label{subsec:metrics-supp}

We evaluate our results quantitatively using three standard metrics defined by Eigen \etal \cite{Eigen2014}.  The first is the root mean square error (RMSE) between the ground-truth depth map $Z$ and the reconstructed depth map $\hat{Z}$, defined below for an $m\times n$ depth map.

\begin{equation}
\mathrm{RMSE} = 
\sqrt{\frac{1}{mn} \sum_{i=1}^{m} \sum_{j=1}^{n} 
\bigl( \hat{Z}_{ij} - Z_{ij} \bigr)^{2}}
\end{equation}

The second metric is the absolute relative error (AbsRel), which measures the average absolute difference between $Z$ and $\hat{Z}$, normalized by $Z$.

\begin{equation}
\mathrm{AbsRel} =
\frac{1}{mn} 
\sum_{i=1}^{m} \sum_{j=1}^{n}
\frac{ \left| \hat{Z}_{ij} - Z_{ij} \right| }{ Z_{ij} }
\end{equation}
    
Finally, we report the $\delta_k$ accuracy metrics, which measure the percentage of pixels where the predicted depth is within a factor of $1.25^k$ of the ground-truth depth.  For $k=1, 2, 3$, we compute:

\begin{equation}
\delta_k 
= \frac{1}{N} 
\sum_{i=1}^{N}
\begin{cases}
1, & \text{if } \max\!\left(
\dfrac{\hat{z}_i}{z_i},\,
\dfrac{z_i}{\hat{z}_i}
\right) < 1.25^{\,k} \\[6pt]
0, & \text{otherwise}
\end{cases}
\end{equation}

\subsection{Extended qualitative results on NYUv2 and Make3D}
\label{subsec:qual}

\begin{figure*}
\begin{minipage}[c][\textheight][c]{\textwidth}
  \centering
   \includegraphics[width=\textwidth]{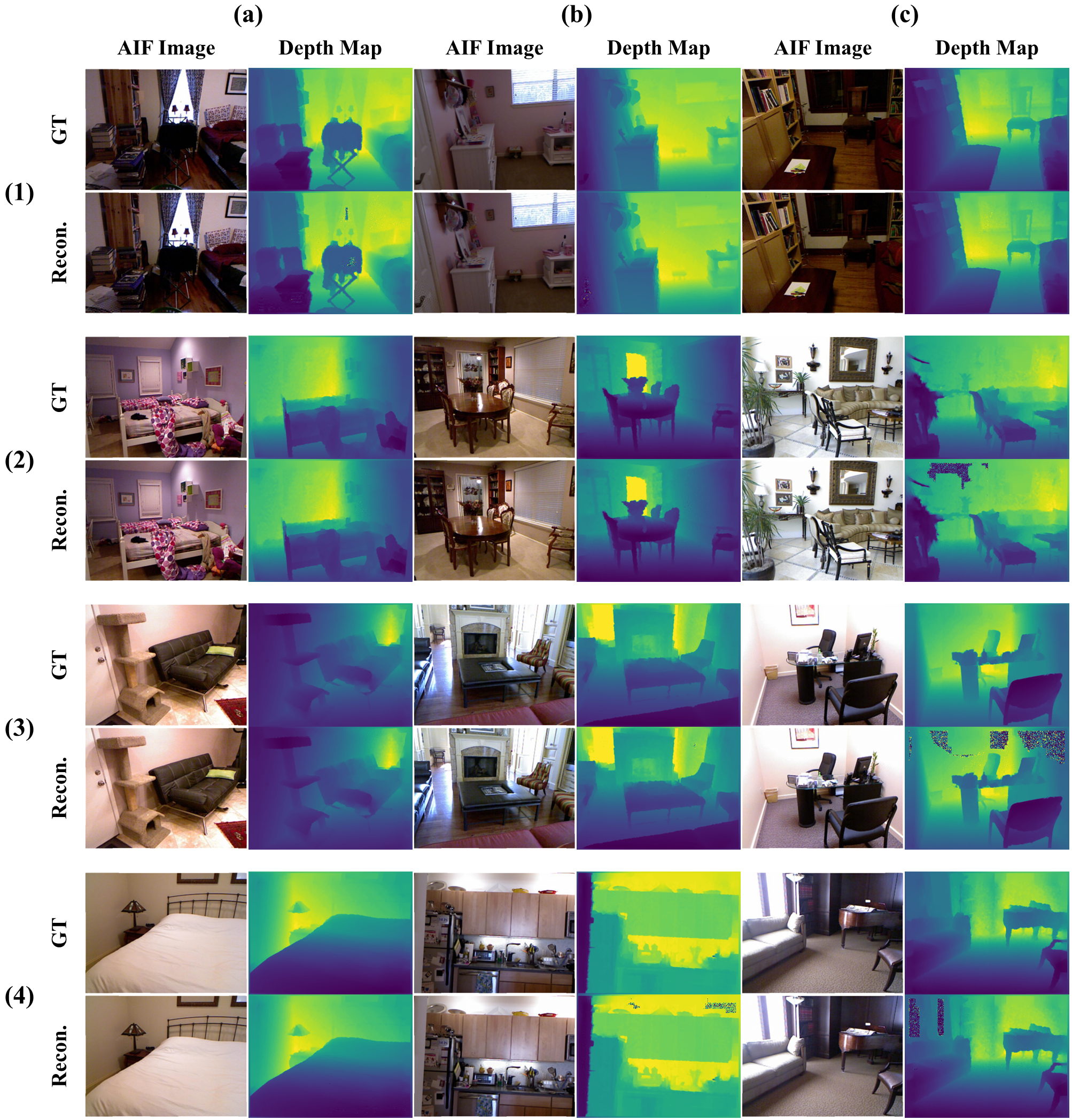}
  \caption{Selection of qualitative results from the NYUv2 dataset.  The top half of the examples (1 and 2) are from the test set, the bottom half of the examples (3 and 4) are from the train set.  First row of each block shows the ground truth all-in-focus images and depth maps, and the second row of each block shows the reconstructed all-in-focus images and depth maps..}
  \label{fig:nyuv2-supp-examples}
\end{minipage}
\end{figure*}

\noindent \textbf{NYUv2 \cite{Silberman2012}.}  Figure \ref{fig:nyuv2-supp-examples} shows extended qualitative results on a variety of examples from both the training and test sets in the NYUv2 dataset \cite{Silberman2012}.  Figure \ref{fig:convergence-nyuv2} shows the corresponding convergence curves under the main-paper settings.

In contrast to approaches such as Si \etal \cite{Si2023}, we capture finer details and avoid the oversmoothing typically seen in prior work.  Nonetheless, the method presents some recognizable artifacts that appear in predictable areas.  For instance, in bright and oversaturated areas that lack texture or color variation across a large patch, our method produces artifacts -- such as those visible in 1a and 4c (e.g. near the window) and in 2c (on the uniform white wall) in Figure \ref{fig:nyuv2-supp-examples}.  However, these issues arise only in particularly tricky cases. For instance in 4a of Figure \ref{fig:nyuv2-supp-examples} (the white bed), our method still produces an accurate reconstruction despite the minimal texture (RMSE = 0.00383).

We also observe that the AIF reconstruction exhibits remarkably high fidelity, despite the fact that the input focal stacks lacked in-focus regions for several parts of the scene.\medskip

\begin{figure*}
  \centering
   \resizebox{\textwidth}{0.95\textheight}{\includegraphics{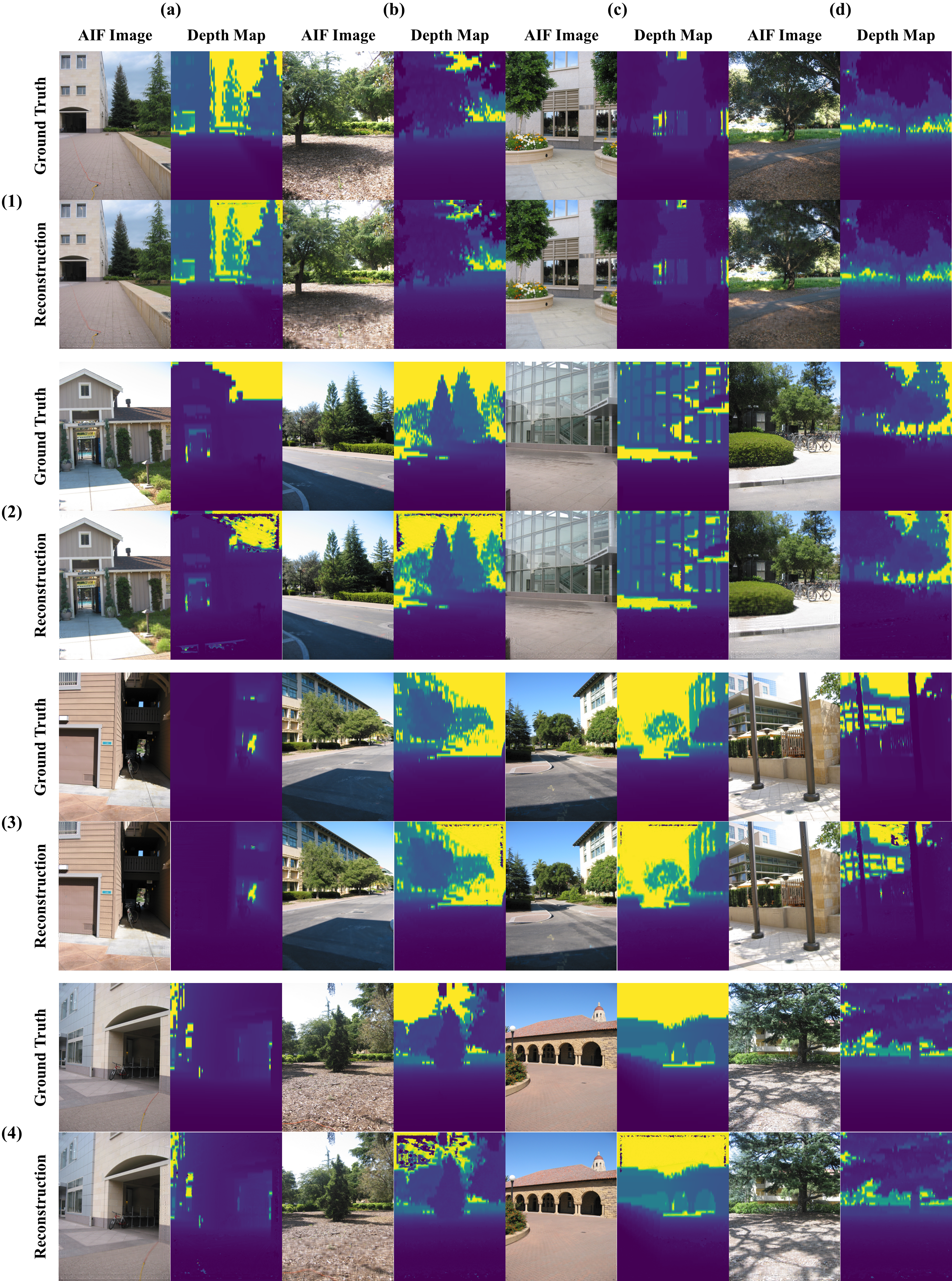}}
  \caption{Selection of qualitative results from the Make3D dataset.  The top half of the examples (1 and 2) are from the test set, the bottom half of the examples (3 and 4) are from the train set.  First row of each block shows the ground truth all-in-focus images and depth maps, and the second row of each block shows the reconstructed all-in-focus images and depth maps.}
  \label{fig:make3d-supp-examples}
\end{figure*}

\noindent \textbf{Make3D \cite{Saxena2005, Saxena2009}.} Figure \ref{fig:make3d-supp-examples} presents extended qualitative results on a variety of examples from both the training and test sets of the Make3D dataset \cite{Saxena2005, Saxena2009}.   Figure \ref{fig:convergence-make3d} shows the corresponding convergence curves under the main-paper settings.

Similar to the NYUv2 results, the method preserves fine details.  This is visually evident in the reconstruction of trees and architecture in Figure \ref{fig:make3d-supp-examples}, for example.  The Make3D results also suffer from artifacts in areas with limited scene texture, such as homogeneous bright sky regions (e.g. 4b in Figure \ref{fig:make3d-supp-examples}). However, even slight variation in color or texture is sufficient for stable reconstruction (as seen in 1a in Figure \ref{fig:make3d-supp-examples} with light cloud structure).

The AIF reconstructions on Make3D are not quite as visually accurate as those on NYUv2, but they were generated with significantly fewer iterations, resulting in a substantial speed-up. This reflects an inherent tradeoff. Even so, the reconstructions are still reasonable considering that some regions in the input Make3D focal stacks are completely blurred.

\subsection{Comparisons to monocular depth estimation methods}
\label{subsec:mono-depth-comp}

Monocular depth estimation approaches predict depth from a single (often defocused) image.  The majority of popular monocular depth estimation methods are fully supervised and trained directly on ground-truth depth maps, including Li \etal \cite{Li2015}, Xu \etal \cite{Xu2017}, Anwar \etal \cite{Anwar2017}, Fu \etal \cite{Fu2018}, Tang \etal \cite{Tang2021}, Dong \etal \cite{Dong2022}, and Nazir \etal \cite{Nazir2023}.  Other approaches such as Zhou \etal \cite{Zhou2011}, Godard \etal \cite{Godard2017}, and Wang \etal \cite{Wang2018} instead use stereo or monocular video supervision in place of ground-truth depth.

A small number of methods deliberately integrate optical models of defocus into their formulation.  Gur and Wolf \cite{GurandWolf2019} train under supervision from the ground-truth AIF images and explicitly incorporate the depth-from-defocus forward model into their training pipeline.  Chang and Wetzstein \cite{ChangandWetzstein2019} jointly optimize the optical system (including lens and input image blur) alongside a fully supervised depth-estimation network.

Table \ref{tab:NYUv2_results_supp} compares state-of-the-art depth-from-defocus and monocular depth estimation methods on the NYUv2 dataset.  Table \ref{tab:Make3D_results_supp} compares state-of-the-art depth-from-defocus and monocular depth estimation methods on the Make3D dataset.   Across both the NYUv2 and Make3D datasets, our method outperforms all of the aforementioned methods with recorded results, except for the $\delta_3$ accuracy metric achieved by Chang and Wetzstein \cite{ChangandWetzstein2019}, whose approach benefits from an algorithmically-optimized lens design.

These results show that our method is not only strong within the depth-from-defocus paradigm, but also competitive with a wide range of approaches that tackle single-view depth reconstruction using different cues and assumptions.

\begin{table}[t]
\centering
\setlength{\tabcolsep}{3pt} 
\small
\begin{tabular}{lccccc}
\toprule
Method & RMSE$\downarrow$ & AbsRel$\downarrow$ & $\delta_1\uparrow$ & $\delta_2\uparrow$ & $\delta_3\uparrow$\\
\midrule
\multicolumn{6}{c}{\textit{Supervised learning (DfD)}} \\
{Song \& Lee \cite{SongandLee2018}} & 0.154 & 0.028 & - & - & - \\
{Carvalho \etal \cite{Carvalho2018}} & 0.144 & 0.036 & - & - & - \\
{Yang \etal (DFF-FV) \cite{Yang2022}}    & 0.285 & 0.470 & 0.956 & 0.979 & 0.988 \\
{Yang \etal (DFF-DFV) \cite{Yang2022}}   & 0.232 & 0.445 & 0.967 & 0.980 & 0.990 \\
\midrule
\multicolumn{6}{c}{\textit{Supervised learning (mono. depth est.)}} \\
{Li \etal \cite{Li2015}} & 0.635 & 0.143 & 0.788 & 0.958 & 0.991 \\
{Anwar \etal \cite{Anwar2017}} & 0.347 & 0.094 & - & - & - \\
{Xu \etal (MS-CRF) \cite{Xu2017}} & 0.586 & 0.121 & 0.811 & 0.954 & 0.987 \\
{Fu \etal (DORN) \cite{Fu2018}} & 0.509 & 0.115 & 0.828 & 0.965 & 0.992 \\
{Chang \& Wetzstein \cite{ChangandWetzstein2019}} & 0.433 & 0.087 & 0.930 & 0.990 & \textbf{0.999} \\
{Tang \etal \cite{Tang2021}} & 0.579 & 0.132 & 0.826 & 0.936 & 0.992 \\
{Dong \etal \cite{Dong2022}} & 0.537 & 0.146 & 0.799 & 0.951 & 0.988 \\
{Nazir \etal \cite{Nazir2023}} & 0.244 & 0.029 & 0.914 & 0.979 & 0.995 \\
\midrule
\multicolumn{6}{c}{\textit{Self-supervised w/ AIF}} \\
{Gur \& Wolf \cite{GurandWolf2019}} & 0.649 & 0.184 & 0.720 & 0.887 & 0.951 \\
{Lu \etal (Defocus-Net) \cite{Lu2021}} & 0.623 & 0.176 & 0.732 & 0.887 & 0.951 \\
{Lu \etal (Focus-Net) \cite{Lu2021}} & 0.611 & 0.172 & 0.748 & 0.892 & 0.949 \\
\midrule
\multicolumn{6}{c}{\textit{Self-supervised}} \\
{Si \etal \cite{Si2023}} & 0.325 & 0.170 & 0.950 & 0.979 & 0.987 \\
\midrule
\multicolumn{6}{c}{\textit{Analytical}} \\
{Moeller \etal \cite{Moeller2015}} & 0.985 & 0.263 & 0.670 & 0.778 & 0.912 \\
{Suwajanakorn \etal \cite{Suwajanakorn2015}} & 0.950 & 0.250 & 0.688 & 0.802 & 0.917 \\
{Ours} & \textbf{0.109} & \textbf{8.37e-3} & \textbf{0.992} & \textbf{0.994} & 0.995 \\
\bottomrule
\end{tabular}
\caption{Extended comparison of depth-from-defocus and monocular depth estimation results on the NYUv2 test set \cite{Silberman2012}. Results for \cite{Moeller2015, Suwajanakorn2015, GurandWolf2019, Lu2021} are recorded from \cite{Lu2021}; results for \cite{Si2023, Yang2022} from \cite{Si2023}; results for \cite{Li2015, Xu2017, Fu2018} from \cite{GurandWolf2019}; and results for \cite{Anwar2017, Tang2021, ChangandWetzstein2019, Dong2022, SongandLee2018, Carvalho2018, Nazir2023} from \cite{Nazir2023}. Lower is better for RMSE and AbsRel; higher is better for $\delta_1$, $\delta_2$, and $\delta_3$.}
\label{tab:NYUv2_results_supp}
\end{table}

\begin{table}[t]
\centering
\setlength{\tabcolsep}{3pt} 
\small
\begin{tabular}{lcccc}
\toprule
& \multicolumn{2}{c}{C1} & \multicolumn{2}{c}{C2}\\
Method & RMSE$\downarrow$ & AbsRel$\downarrow$ & RMSE$\downarrow$ & AbsRel$\downarrow$\\
\midrule
\multicolumn{5}{c}{\textit{Supervised learning}} \\
{Li \etal \cite{Li2015}} & 7.120 & 0.278 & 10.27 & 0.279 \\
{Xu \etal (MS-CRF) \cite{Xu2017}} & 4.380 & 0.184 & 8.56 & 0.198 \\
{Fu \etal (DORN) \cite{Fu2018}} & 3.970 & 0.157 & 7.32 & 0.162 \\
{Nazir \etal \cite{Nazir2023}} & 4.178 & 0.153 & 6.132 & 0.170 \\
\midrule
\multicolumn{5}{c}{\textit{Self-supervised learning with stereo/monocular video}} \\
{Godard \etal \cite{Godard2017}} & 11.513 & 0.443 & - & - \\
{Zhou \etal \cite{Zhou2017}} & 10.470 & 0.383 & - & - \\
{Wang \etal \cite{Wang2018}} & 8.090 & 0.387 & - & - \\
\midrule
\multicolumn{5}{c}{\textit{Self-supervised with AIF}} \\
{Gur \& Wolf \cite{GurandWolf2019}} & 7.474 & 0.262 & 9.248 & 0.269 \\
\midrule
\multicolumn{5}{c}{\textit{Analytical}} \\
{Ours} & \textbf{2.277} & \textbf{0.0663} & \textbf{5.630} & \textbf{0.0695} \\
\bottomrule
\end{tabular}
\caption{Extended comparison of depth-from-defocus and monocular depth estimation methods on the Make3D test set \cite{Saxena2005, Saxena2009}. Results for \cite{Li2015,GurandWolf2019,Xu2017,Fu2018,Godard2017,Zhou2017,Wang2018} are recorded from \cite{GurandWolf2019}. For Gur and Wolf \cite{GurandWolf2019}, we report their best RMSE result (F6, rendered focal stack with six focus settings). Lower is better.}
\label{tab:Make3D_results_supp}
\end{table}

\subsection{Complete results on mobile phone focal stacks}
\label{subsec:mobile-depth-supp}

We evaluate our method on the 11 mobile-phone focal stacks from Suwajanakorn \etal \cite{Suwajanakorn2015}. In the main paper, we present three representative examples; the same ones are used for comparison in Figure 5 of Si \etal \cite{Si2023}.  Below, we include the full set of results for all 11 stacks.

\begin{figure}[t]
  \centering
   \includegraphics[width=\linewidth]{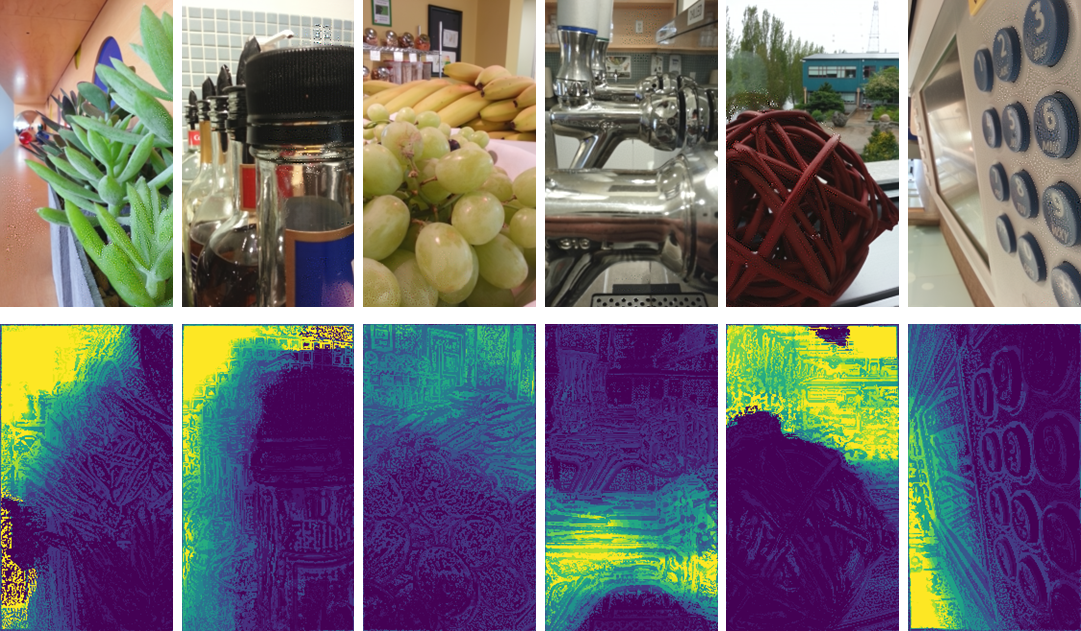}
  \caption{Reconstructed all-in-focus images (top row) and depth maps (bottom row) for a selection of (vertically-oriented) mobile phone focal stacks from Suwajanakorn \etal \cite{Suwajanakorn2015}.}
  \label{fig:mobile-depth-vertical}
\end{figure}

\begin{figure}[t]
  \centering
   \includegraphics[width=\linewidth]{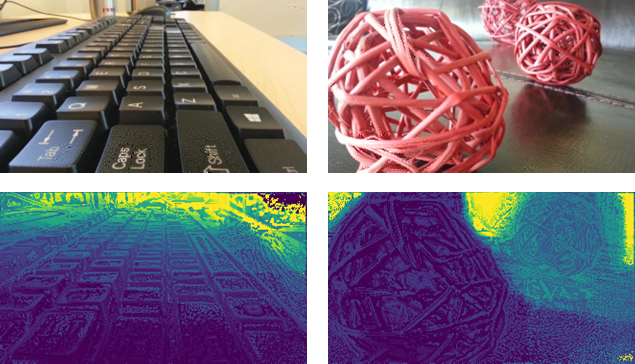}
  \caption{Reconstructed all-in-focus images (top row) and depth maps (bottom row) for two (horizontally-oriented) mobile phone focal stacks (of a keyboard and rubber band balls) from Suwajanakorn \etal \cite{Suwajanakorn2015}.}
  \label{fig:mobile-depth-horizontal}
\end{figure}

\begin{figure}[t]
  \centering
   \includegraphics[width=\linewidth]{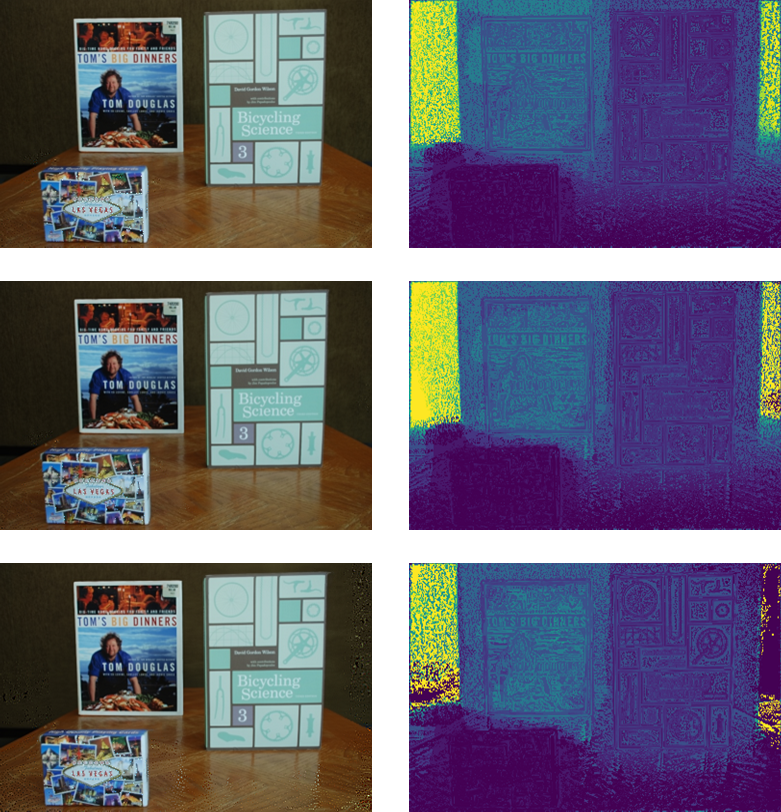}
  \caption{Reconstructed all-in-focus images (left column) and depth maps (right column) of three (aligned) focal stacks from Suwajanakorn \etal \cite{Suwajanakorn2015}.}
  \label{fig:mobile-depth-motion}
\end{figure}

Figure \ref{fig:mobile-depth-vertical} highlights the key examples shown in Figure 5 of Suwajanakorn \etal \cite{Suwajanakorn2015}.  Figure \ref{fig:mobile-depth-horizontal} shows our results on the ball and keyboard examples, corresponding to the reconstructions in Figures 1 and 7 of Suwajanakorn \etal \cite{Suwajanakorn2015}. Figure \ref{fig:mobile-depth-motion} presents results on three examples that were originally captured with different amounts of camera motion; however, we use the aligned focal stacks provided by the authors. The corresponding comparison for this example appears in their Figure 6 \cite{Suwajanakorn2015}.

\begin{table*}[!t]
\centering
\setlength{\tabcolsep}{3pt} 
\small
\begin{tabular}{ccc|ccccc|ccccc}
\toprule
\multicolumn{3}{c|}{Settings} & \multicolumn{5}{c|}{Train Set} & \multicolumn{5}{c}{Test Set}\\
$T_0$ & $\alpha$ & \# epochs
& RMSE$\downarrow$ & AbsRel$\downarrow$ & $\delta_1\uparrow$ & $\delta_2\uparrow$ & $\delta_3\uparrow$
& RMSE$\downarrow$ & AbsRel$\downarrow$ & $\delta_1\uparrow$ & $\delta_2\uparrow$ & $\delta_3\uparrow$\\
\midrule
200 & 1.05 & 40
& 0.110 & 8.11e-3 & 0.992 & 0.994 & 0.995
& 0.109 & 8.37e-3 & 0.992 & 0.994 & 0.995 \\
50 & 2 & 5
& 0.189 & 0.0198 & 0.985 & 0.990 & 0.992
& 0.190 & 0.0211 & 0.984 & 0.989 & 0.991 \\
10 & 2 & 5
& 0.348 & 0.0394 & 0.964 & 0.981 & 0.985
& 0.333 & 0.0386 & 0.967 & 0.982 & 0.986 \\
\bottomrule
\end{tabular}
\caption{Performance of our method on NYUv2 under different parameter settings. $T_0$ is the initial maximum number of iterations for the Nesterov gradient method; $\alpha$ is the multiplier that increases the number of Nesterov iterations at each alternating-minimization step; and \# epochs is the number of alternating-minimization epochs.}
\label{tab:NYUv2_settings}
\end{table*}

\begin{table*}[!t]
\centering
\setlength{\tabcolsep}{3pt} 
\small
\begin{tabular}{ccccc|cccc|cccc}
\toprule
\multicolumn{5}{c|}{Settings} & \multicolumn{4}{c|}{Train Set} & \multicolumn{4}{c}{Test Set}\\
\multicolumn{3}{c}{} & \multicolumn{2}{c|}{Windowed MSE} & \multicolumn{2}{c}{C1} & \multicolumn{2}{c|}{C2} & \multicolumn{2}{c}{C1} & \multicolumn{2}{c}{C2}\\
$T_0$ & $\alpha$ & \#epochs & Enabled? & Size
& RMSE$\downarrow$ & AbsRel$\downarrow$ & RMSE$\downarrow$ & AbsRel$\downarrow$
& RMSE$\downarrow$ & AbsRel$\downarrow$ & RMSE$\downarrow$ & AbsRel$\downarrow$\\
\midrule
10 & 2 & 5 & No & --
& 2.487 & 0.0850 & 7.302 & 0.0971
& 2.432 & 0.0751 & 7.924 & 0.0881 \\
10 & 2 & 5 & Yes & 5
& 2.405 & 0.0781 & 5.729 & 0.0824
& 2.277 & 0.0663 & 5.630 & 0.0695 \\
\bottomrule
\end{tabular}
\caption{Effect of windowed MSE on optimization over Make3D dataset. We report train and test performance under the C1/C2 evaluation protocols. When enabled, windowed MSE averages the per-pixel reconstruction error over a local window of a user-defined size, improving accuracy.}
\label{tab:Make3D_settings}
\end{table*}

\section{Performance and parameter tradeoffs}
\label{sec:supp-performance}

This section presents additional experiments showing how performance changes as we vary a few key parameters. We first vary the main runtime controls (the FISTA iteration budget and the number of alternating-minimization iterations) and show how the resulting performance changes on the NYUv2 dataset (Section \ref{sec:tradeoffs}). We then show the effect of using a windowed MSE during grid search on the Make3D dataset (Section \ref{sec:window-exp}).

\subsection{Effect of iteration count on NYUv2 performance}\label{sec:tradeoffs}

The primary factors influencing runtime are the grid-search sample count $n$, the initial FISTA iteration limit $T_0$, the FISTA iteration-scaling parameter $\alpha$, and the total number of epochs in the alternating-minimization procedure.  These parameters were defined in Section 3 the main paper.  Because the grid search is accelerated using convolution, the number of grid-search samples $n$ has the least impact on runtime.  In this section, we examine the tradeoffs between $T_0$, $\alpha$, and the number of alternating-minimization epochs on the NYUv2 dataset.

Table \ref{tab:NYUv2_settings} reports performance under three different parameter settings.  Even when operating under more permissive tolerance settings ($T_0 = 10$, $\alpha = 2$, and 5 alternating-minimization epochs), our method remains competitive with the unsupervised neural network of Si \etal \cite{Si2023}, while still outperforming several fully-supervised monocular depth-estimation methods, approaches that are supervised on the AIF images, and the analytical baselines.  At $T_0 = 50$, $\alpha = 2$, and 5 epochs, we outperform all methods (in terms of RMSE and AbsRel) except the supervised DfD models of Song \& Lee \cite{SongandLee2018} and Carvalho \etal \cite{Carvalho2018}.  Using 
$T_0=200$, $\alpha = 1.05$, and 40 epochs yields a considerably more conservative and computationally expensive configuration, yet it delivers the strongest performance overall -- with the sole exception of the $\delta_3$ metric reported by Chang and Wetzstein \cite{ChangandWetzstein2019}.  These settings were used to achieve the results in the main paper on the NYUv2 dataset.  However, more relaxed tolerances of $T_0 = 10$, $\alpha = 2$, and 5 epochs were used to achieve competitive results on the Make3D data in the main paper.

\subsection{Effect of windowed MSE during grid search on Make3D performance}\label{sec:window-exp}

In addition, we analyze the effect of windowed MSE during grid search on the Make3D results. In the main experiments (Sec. 4), we enabled windowed MSE during grid search on Make3D and the mobile phone focal stacks, because it slightly improved performance on Make3D and significantly improved the qualitative results on real blur.  However, it was not necessary to obtain competitive results on NYUv2. Table \ref{tab:Make3D_settings} compares train and test performance with and without windowed MSE on Make3D. Without windowed MSE, our method is already on par with (or better than) all methods reported in Table \ref{tab:Make3D_results_supp}; enabling windowed MSE further improves accuracy and yields the best overall performance.

\vfill\pagebreak


\noindent\begin{center}
\bfseries\MakeUppercase{SUPPLEMENTARY REFERENCES}
\end{center}

\makeatletter
\renewcommand{\refname}{\vspace*{-1.8\baselineskip}}
\makeatother

{\small
\bibliographystyle{IEEEbib}
\bibliography{supp,refs}
}